\documentclass[10pt,twocolumn,letterpaper]{article}

\usepackage{iccv}
\usepackage{times}
\usepackage{epsfig}
\usepackage{graphicx}
\usepackage{amsmath}
\usepackage{amssymb}

\usepackage[accsupp]{axessibility}
\usepackage{booktabs}
\usepackage{tabularx}

\newcommand{\Fref}[1]{Fig.~\ref{#1}}
\newcommand{\Sref}[1]{Sec.~\ref{#1}}
\newcommand{\Tref}[1]{Tab.~\ref{#1}}

\newcommand{\Eref}[1]{Eq.~\ref{#1}}

\usepackage{capt-of}
\usepackage{lipsum}

\newcommand\blfootnote[1]{%
  \begingroup
  \renewcommand\thefootnote{}\footnote{#1}%
  \addtocounter{footnote}{-1}%
  \endgroup
}

\usepackage[pagebackref=true,breaklinks=true,letterpaper=true,colorlinks,bookmarks=false]{hyperref}

\iccvfinalcopy 

\def\iccvPaperID{****} 

\ificcvfinal\fi

\begin{document}

\title{AesPA-Net: Aesthetic Pattern-Aware Style Transfer Networks}

\author{Kibeom Hong \textsuperscript{1,3,4}\qquad Seogkyu Jeon\textsuperscript{1}\qquad Junsoo Lee\textsuperscript{4}\qquad Namhyuk Ahn\textsuperscript{4}\qquad Kunhee Kim\textsuperscript{2,4}\\Pilhyeon Lee\textsuperscript{1}\qquad Daesik Kim\textsuperscript{4}\qquad Youngjung Uh\textsuperscript{1}\qquad Hyeran Byun\textsuperscript{1*}\vspace{1.5mm}\\
\textsuperscript{1}Yonsei University\qquad\textsuperscript{2}KAIST AI\qquad \textsuperscript{3}SwatchOn\qquad\textsuperscript{4}NAVER WEBTOON AI
}

\ificcvfinal\thispagestyle{empty}\fi

\twocolumn[{
\renewcommand\twocolumn[1][]{#1}
\maketitle
\begin{center}
    \centering
    \includegraphics[width=1.0\linewidth]{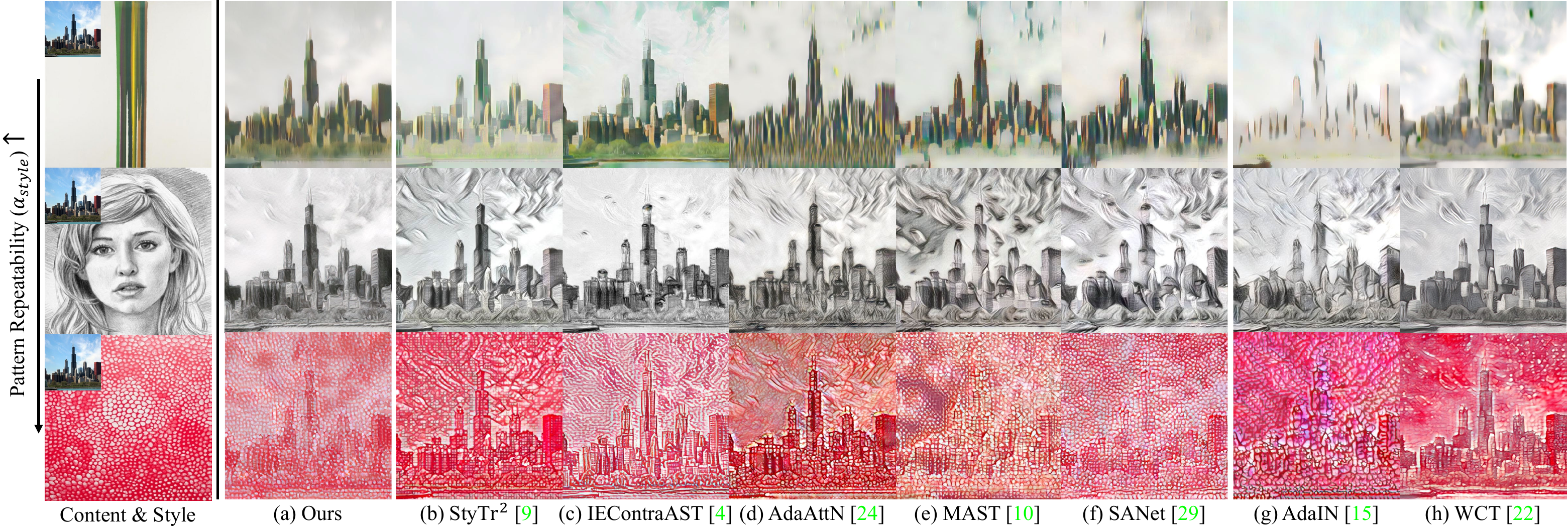}
    \captionof{figure}{Stylization results. Given input pairs~(a fixed content and three references), we compare our method (a) with both attention-based transfer methods (b-f) and global statistic-based methods (g-h). While previous works fail to generate plausible results with various style, our model successfully conducts artistic style transfer well regardless of different style pattern. (Best viewed in zoomed and color.)}
    \label{fig:teaser}
\end{center}
}]
\blfootnote{* Corresponding author}

\begin{abstract}
   To deliver the artistic expression of the target style, recent studies exploit the attention mechanism owing to its ability to map the local patches of the style image to the corresponding patches of the content image. However, because of the low semantic correspondence between arbitrary content and artworks, the attention module repeatedly abuses specific local patches from the style image, resulting in disharmonious and evident repetitive artifacts. To overcome this limitation and accomplish impeccable artistic style transfer, we focus on enhancing the attention mechanism and capturing the rhythm of patterns that organize the style. In this paper, we introduce a novel metric, namely pattern repeatability, that quantifies the repetition of patterns in the style image. Based on the pattern repeatability, we propose \textbf{Aes}thetic \textbf{P}attern-\textbf{A}ware style transfer Networks (AesPA-Net) that discover the sweet spot of local and global style expressions. In addition, we propose a novel self-supervisory task to encourage the attention mechanism to learn precise and meaningful semantic correspondence. Lastly, we introduce the patch-wise style loss to transfer the elaborate rhythm of local patterns. Through qualitative and quantitative evaluations, we verify the reliability of the proposed pattern repeatability that aligns with human perception, and demonstrate the superiority of the proposed framework. All codes and pre-trained weights are available at
{\small \href{https://github.com/Kibeom-Hong/AesPA-Net}{Kibeom-Hong/AesPA-Net}}.
\end{abstract}

\section{Introduction}
Style transfer aims to render the content of a source image using the elements of a style image. Although the concepts of content and style cannot be rigorously defined, the research community has been building agreeable ideas, considering repeating elements such as brush strokes, color maps, patterns, and certain dominant shapes to be style \cite{gatys2016image}. Since Gatys et al. \cite{gatys2016image} pioneered the Neural Style Transfer by using the Gram matrix of deep feature activations as the style representation, previous works have shown surprising performance in Artistic Style Transfer (AST) by employing the global transformation methods \cite{li2017universal,huang2017arbitrary,An2021ArtFlowUI} and patch-wise swapping approaches \cite{Chen2016Fast,sheng2018avatar,jing2022learning} that leverage the similarity between patches in the content and style inputs. However, a major challenge for these methods is the lack of semantic information on the content images, which often leads to distortion artifacts (\Fref{fig:teaser} (g-h)).

To handle this, recent advances in AST methods \cite{park2019arbitrary,Deng2020ArbitraryST,Liu2021AdaAttNRA,Luo2022ConsistentST,deng2022stytr2,Zhang2022DomainEA} have incorporated the attention mechanism which aggregates elements from a style image to render the content of a source image according to semantic correspondence between local patches of the images. While the principle of attention networks has shown the great potential in guiding detailed expression and maintaining content information, they often produce abnormal patterns or evident artifacts, \eg{, smudges or floating eyes} (\Fref{fig:teaser} (d-f)). Although several works \cite{deng2022stytr2,Chen2021ArtisticST,wang2022aesust} have attempted to alleviate this problem by employing transformer or external learning strategies, achieving detailed textures such as pointillism or brush strokes remains challenging (\Fref{fig:teaser} (b-c)).

In order to overcome above issues, we conjecture pitfalls of the attention mechanism for AST. Firstly, we point out that the low semantic correspondence between arbitrary content and style images induces attention mechanisms to focus on limited regions of the style image. This can hinder attention-based methods from accurately capturing and expressing the entire style of the reference images. Consequently, this obstacle leads to disharmonious artifacts since they heavily rely on few style patches such as eyes instead of a representative style \eg, a pencil sketch in a portrait ($2^{nd}$ row of \Fref{fig:teaser}). Furthermore, attention mechanisms utilize only small patches from the style image where large patches would be more suitable. This leads to overly repetitive patterns, even with less-repetitive styles. For instance, despite a simple stripe style, global regions such as the sky are prone to contain artifacts and objects can be painted with excessively repetitive stripes ($1^{st}$ row of \Fref{fig:teaser}).

\begin{figure}[t]
        \centering
        \includegraphics[width=1.00\linewidth]{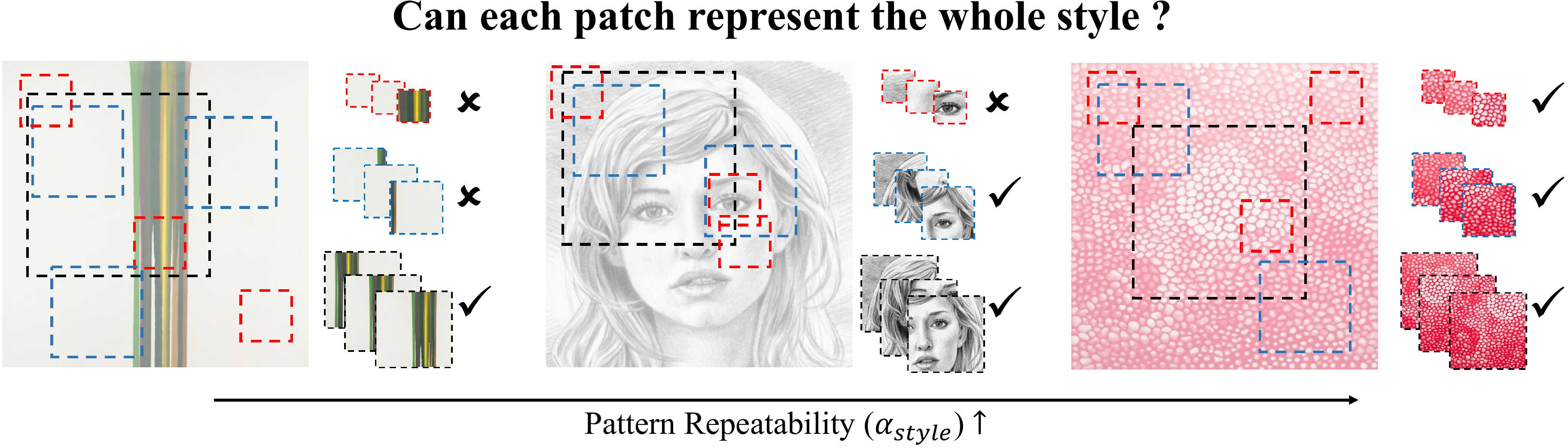}
        \captionof{figure}{Illustration of our motivation. We show whether each local pattern with various scale could represent the entire style images or not. Motivated by this, we introduce a new metric to quantify the repetition of local patches, \ie{, pattern repeatability}.}
        \label{fig:motivation}
\end{figure}

To this end, we introduce remedies for more delicate artistic expression and improving the attention mechanism. First, we revisit the definition of style considering its unique repeatability. As shown in \Fref{fig:motivation}, our motivation is based on the observation that every style can be expressed as a repetition of appropriate patches which could describe the entire style. To this end, we propose a novel metric \emph{pattern repeatability} which quantifies the frequency of repeating patches in a style image. Then it indicates whether we should bring more effects from attention-based stylization whose advantage lies in details, or global statistic-based stylization whose advantage lies in the smooth reflection of entire styles. Accordingly, we introduce the \textbf{Aes}thetic \textbf{P}attern-\textbf{A}ware style transfer Networks (\textit{AesPA-Net}) which calibrate the stylized features from the two approaches upon the pattern repeatability.

In addition, we propose the self-supervisory task for encouraging the attention module to capture the broader corresponding regions of style images even for arbitrary pairs. This auxiliary task amounts to maximizing the similarity between the augmented input and its original at the feature level. As a result, it effectively reinforces the attention modules to conduct the artistic stylization. Last but not least, we modify the style loss to be computed between patches with the proper size. Our patch-wise style loss induces the stylized results to reflect the rhythm of local patterns according to the proposed pattern repeatability.

In experiments, we qualitatively show that our networks are capable of articulate artistic stylization with various patterns of style. Besides, through quantitative proxy metrics and the user study, we demonstrate that our framework, \textit{AesPA-Net}, outperforms the state-of-the-art AST studies: five attention-based methods \cite{park2019arbitrary,Deng2020ArbitraryST,Liu2021AdaAttNRA,Chen2021ArtisticST,deng2022stytr2} and two global statistics-based methods \cite{li2017universal,huang2017arbitrary}. Furthermore, we verify that the reliability of pattern repeatability closely aligns with human perception, and provide ablation studies to validate the effect of individual components.

\section{Related works}
\label{sec:related}
\subsection{Artistic Style Transfer}
\noindent\textbf{Global transformation and patch-wise swapping-based.}
Since Gatys et al. \cite{gatys2016image} opened up the research of Neural Style Transfer (NST), NST studies have been explored in various domains such as photo-realistic \cite{yoo2019photorealistic,Xia2020JointBL,An2020UltrafastPS}, video \cite{Huang2017RealTimeNS,Wang2020ConsistentVS,Chen2020OpticalFD,Deng2021ArbitraryVS}, and 3D \cite{Hllein2022StyleMeshST,Mu20223DPS}. Among others, Artistic Style Transfer (AST) \cite{li2017universal,huang2017arbitrary,Chen2016Fast,sheng2018avatar} has drawn huge and broad attention by sublimating daily photos into artistic expressions. Recently, Li~\etal~\cite{li2017universal} and Huang~\etal~\cite{huang2017arbitrary} introduce the Whitening and Coloring Transformation~(WCT) and the Adaptive Instance Normalization~(AdaIN) to achieve arbitrary style transfer by washing away the global feature statistics and replacing them with the target ones. Meanwhile, Chen \etal \cite{Chen2016Fast} swap the patches of content features and Sheng \etal \cite{sheng2018avatar} propose the style decorator to normalize feature with the most correlated style features. Jing \etal \cite{jing2022learning} have improved semantic matching by employing techniques such as Graph Neural Networks. However, these AST methods have difficulty in preserving the original context due to the lack of cues from content images, resulting in unrealistic stylization as well as content distortion. To overcome this limitation, attention-based local patch stylization methods \cite{yao2019attention,park2019arbitrary,Deng2020ArbitraryST,Liu2021AdaAttNRA,Chen2021ArtisticST,Zhang2022DomainEA,wang2022aesust,deng2022stytr2} are proposed.

\noindent\textbf{Attention mechanism-based.}
AAMS \cite{yao2019attention} and SANet \cite{park2019arbitrary} first employ the self-attention mechanism to derive the semantic mapping for artistic style transfer. MAST \cite{Deng2020ArbitraryST} adopts the attention mechanism to not only stylized features but also to content and style features itself. In addition, AdaAttN \cite{Liu2021AdaAttNRA} proposes the fusion strategy that exploits both the attention mechanism and the global feature statistic-based method. Besides, StyTr$^{2}$ \cite{deng2022stytr2} introduces the Transformer and the content-aware positional encoding for delicate style expression. Nonetheless, they suffer from peculiar repetitive artifacts when they confront a style image with irregular patterns, producing abnormal stylization results. IEcontraAST \cite{Chen2021ArtisticST}, CAST \cite{Zhang2022DomainEA}, and AesUST \cite{wang2022aesust} have tackled this problem by exploiting the contrastive and adversarial learning but still fail to express the detailed texture. In this paper, we delve into the relation between the stylization result and the degree of pattern repetition and utilize this relation to better translate the artistic rhythm of the style image, achieving elaborate and aesthetic style transfer.

\subsection{Relation between Pattern and Style}
A pattern appears in various scales in a single image. In other words, a pattern could be described as an entire image or as a very small local patch. Based on this notion, there are a few studies \cite{Julesz1057698,Julesz1984ABO,jing2018stroke,tmm9143296,park2010swappingAE} investigating the correlation between pattern and style. Julez \etal \cite{Julesz1057698,Julesz1984ABO} introduce the texton theory that the human vision system is related to the frequency of repeated patterns. Jing \etal \cite{jing2018stroke} and Virtusio \etal \cite{tmm9143296} propose controllable artistic style transfer networks by manipulating the resolution of reference images. They demonstrate that features extracted from various resolutions of the target image can be interpreted as having different styles, which in turn change the stylization result.
In addition, SwappingAE \cite{park2010swappingAE} proposes the patch discriminator to criticize the texture based on multi-scale patterns and conduct the texture translation. Furthermore, Cheng \etal \cite{Cheng2021StyleAwareNL} statistically analyze the degree of stylization and introduce the normalized style loss to express proper repeatability. Our motivation is also in the continuation of these studies: ``\textit{The unique pattern repeatability in each image defines its own style}". In this paper, we aim to capture not only the elements of the style (\eg{, color or texture}) but also their unique repeatability in organizing the artistic rhythm and expression.

\begin{figure*}[t]
        \centering
        \includegraphics[width=1.0\linewidth]{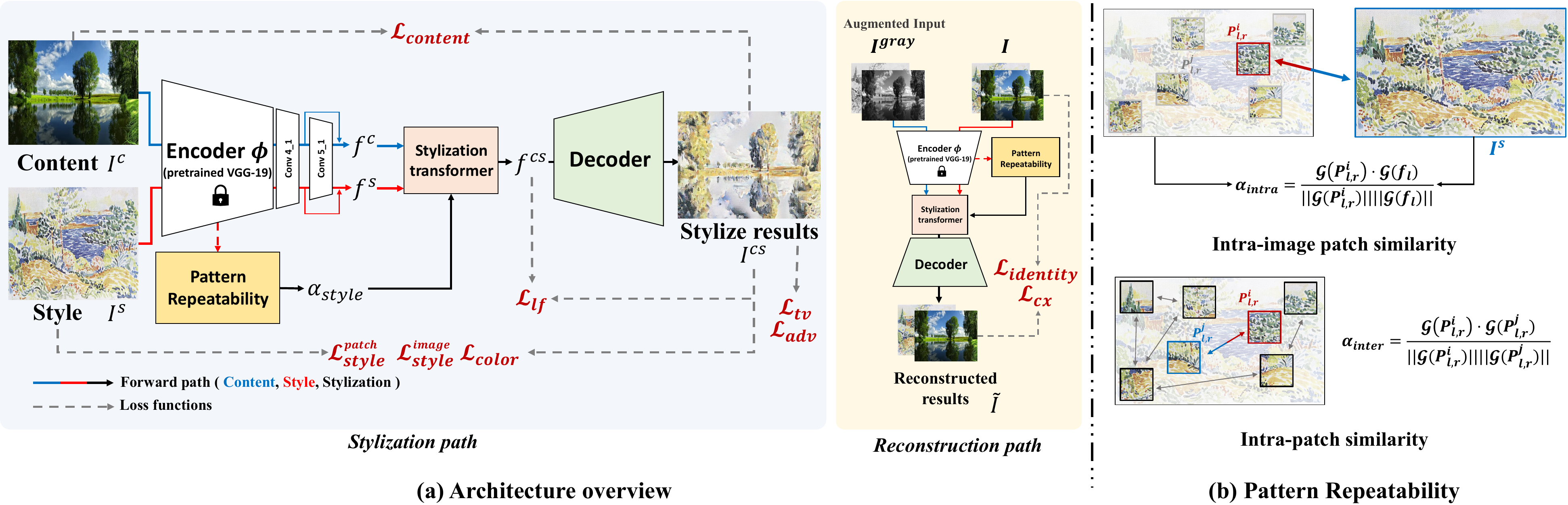}
        \captionof{figure}{The overview of the aesthetic pattern-aware style transfer networks (AesPA-Net). (a) shows the architecture of AesPA-Net and training procedures, \ie{, stylization task (blue box) and reconstruction task (yellow box)}. (b) depicts the detail of pattern repeatability.}
        \label{fig:archi}
\end{figure*}

\section{Proposed Method}
\label{sec:proposed method}
In this section, we describe the proposed \textit{\textbf{pattern repeatability}} and  \textit{\textbf{Aes}thetic \textbf{P}attern-\textbf{A}ware style transfer networks} (AesPA-Net) in detail. Notably, our goal is to render a stylized image ($I^{cs}$) from the content image ($I^{c}$) and the style image ($I^{s}$) by transferring the atmosphere of repeatability as well as elements of style. The overview of our method is illustrated in \Fref{fig:archi}. 

\subsection{Pattern Repeatability}
\label{sec:patternrepeatability}
\noindent \textbf{Motivation.}~Despite the advances of attention-based AST methods, there still exists a fundamental limitation of attention networks in artistic style transfer. Due to the low semantic correspondence between the content and the artistic style images, the attention module tends to map the content to only a few local patches. Hence, a specific local patch pattern of the style image is repeatedly expressed in the stylized results. These stylization results would be visually satisfactory when the small local patches have high similarity to each other as well as the global style, \eg{, pointillism artworks}. Otherwise, the results become unnatural and implausible especially when the local patches cannot represent the overall style, \eg{, portraits}. 
Intrinsically, the degree of local pattern repetition and the minimal patch size, which can represent the whole image, vary across different styles as shown in~\Fref{fig:motivation}. To this end, we propose a novel metric, \textit{pattern repeatability}, to quantify the rhythm and repetition of local patterns in the style image.

\noindent \textbf{Method.}~In order to measure the degree of pattern repeatability in the style image $I^s$, we consider two perspectives: 1) Similarity between each patch and the whole image, \ie{, intra-image patch repeatability}, and 2) Similarity between patches, \ie{, inter-patch repeatability}.

Given an image $I$, we gather features $f_{l} \in \mathbb{R}^{C_{l} \times H_{l} \times W_{l}}$ from the $l^{th}$ convolutional layer of encoder $\phi$, and then obtain $N$ feature patches $\{P^{i}_{l,r}\}_{i=1}^{N}$ by spatially dividing each features with ratio $r$, \ie, $P^{i}_{l,r} \in \mathbb{R}^{C_l \times \frac{H_l}{r} \times \frac{W_l}{r}}$. To estimate how much local styles resemble the global style, \ie{, intra-image patch repeatability $\alpha_{intra}$}, we measure the cosine-similarity between the Gram matrix of each feature patch $P^{i}_{l,r}$ and the entire feature $f_{l}$ as follows.
\begin{equation}
\label{equ:intra}
    {\alpha}_{intra} =  \frac{1}{L}\sum_{l=1}^{L}\frac{1}{N} \sum_{i=1}^{N}\frac{\mathcal{G}(P^{i}_{l,r})\cdot\mathcal{G}(f_{l})}{{\|\mathcal{G}(P^{i}_{l,r})\|}{\left\|\mathcal{G}(f_{l})\right\|}}, 
\end{equation}
where $\mathcal{G}(\cdot)$ depicts the Gram matrix calculation and $L$ indicates the number of convolutional layers in the encoder $\phi$. A high $\alpha_{intra}$ indicates that the global style appears consistently in its local patches, whereas a low $\alpha_{intra}$ means local styles are distinguishable from the global style.

In addition, we calculate the similarity between textures of different patches, \ie{, inter-patch repeatability $\alpha_{inter}$}, to estimate whether particular local patterns appear frequently. This process is formulated as follows.
\begin{equation}
\label{equ:inter}
    {\alpha}_{inter} =  \frac{1}{L}\sum_{l=1}^{L}\frac{1}{|\mathcal{B}|} \sum_{\forall (i,j) \in \mathcal{B}}\frac{\mathcal{G}(P^{i}_{l,r})\cdot\mathcal{G}(P^{j}_{l,r})}{\left\|\mathcal{G}(P^{i}_{l,r})\right\| \left\|\mathcal{G}(P^{j}_{l,r})\right\|}.
\end{equation}
Here $\mathcal{B}$ is the selected pair set out of ${}_{N}{\rm C}_{2}$ pairs with a probability of $p$ for efficient calculation, \ie, $|\mathcal{B}|=p \cdot {}_{N}{\rm C}_{2}$.
Moreover, we compute the pattern repeatability with the grayscaled style image in a similar way to capture the pattern repeatability of the regional structure. Finally, the pattern repeatability is computed as:
\begin{equation}
\label{equ:alpha}
    {\alpha}_{style} =  \frac{1}{4}\cdot({\alpha}^{RGB}_{inter}+{\alpha}^{RGB}_{intra}+{\alpha}^{gray}_{inter}+{\alpha}^{gray}_{intra}).
\end{equation}
We depict the conceptual illustration of pattern repeatability in~\Fref{fig:archi}~(b). Detailed analysis on the reliability of pattern repeatability is provided in~\Sref{sec:analysis}.

\subsection{Aesthetic Pattern-Aware Style Transfer}
\label{sec:AesPA}
We design a novel style transfer networks (AesPA-Net) that can fully reflect the target style by taking its pattern repeatability into account. In order to achieve impeccable artistic stylization with the styles of arbitrary pattern rhythm, the stylization transformer is the key component that adaptively transfers local and global textures in accordance with the pattern repeatability $\alpha_{style}$. Intuitively, the ideal solution for the style with low pattern repeatability would be directly increasing the input patch size of attentional style transfer. Nevertheless, this is impractical because the memory requisition increases quadratic according to the patch scale. As a result, attention maps for stylization are computed at limited levels \ie{, 4th and 5th levels of the encoder}. To overcome this limitation, we implicitly employ a statistics-based transformation as a practical complement to global stylization. Concretely, the global stylized feature $f^{cs}_{global}$ can capture unattended local patterns by matching global statistics to style features of $I^s$. Meanwhile, the features $f^{cs}_{attn}$ from attentional style transfer module emphasize important local patterns and compensates for the distorted content that is caused during whitening.

Consequently, our stylization transformer is composed of an attentional style transfer module and a global feature statistic transformation module. From the patch-wise attentional style transfer module, we obtain the stylized feature $f_{attn}^{cs}$ based on local patch-wise correspondence \cite{Liu2021AdaAttNRA} as:
\begin{equation}
\label{eq1:adaattn}
\begin{split}
    f^{cs}_{l} &= AdaAttN (f^{c}_{l}, f^{s}_{l}, f^{c}_{1:l}, f^{s}_{1:l}), \\
    f^{cs}_{attn} &= h(f^{cs}_{4}+\psi_{\uparrow}(f^{cs}_{5})),
\end{split}
\end{equation}
where $l$ denotes the index of layer, $h(\cdot)$ and $\psi_{\uparrow}$ indicate the convolutional layer and upsampling function, respectively.

To acquire $f^{cs}_{global}$, we apply the global statistic transformation~\cite{li2017universal} on \texttt{conv4\_1} features. After that, we obtain the final stylized feature $f^{cs}$ as:
\begin{equation}
\label{eq1:extract features}
\begin{split}
    f^{cs} &= \alpha_{style} \cdot f_{attn}^{cs} + (1-\alpha_{style}) \cdot f_{global}^{cs},
\end{split}
\end{equation}
where $f^{cs}_{attn}$ and $f^{cs}_{global}$ represent the attended stylized features and global statistic transformed features, respectively. $\alpha_{style}$ is the overall pattern repeatability obtained in \Eref{equ:alpha}.
Finally, our decoder inverts the stylized feature maps to an artistic result $I^{cs}$. In the supplementary material, we show that our stylization transformer also works effectively with another global transformation \ie{, AdaIN~\cite{huang2017arbitrary}}.

As illustrated in~\Fref{fig:archi}~(a), AesPA-Net consists of an encoder, a stylization transformer, and a decoder. We employ the pre-trained VGG-19~\cite{simonyan2014very} as an encoder, while the decoder is designed to mirror the architecture of the encoder by replacing down-sampling operations with up-sampling ones.

\begin{figure}[t]
        \centering
        \includegraphics[width=1.00\linewidth]{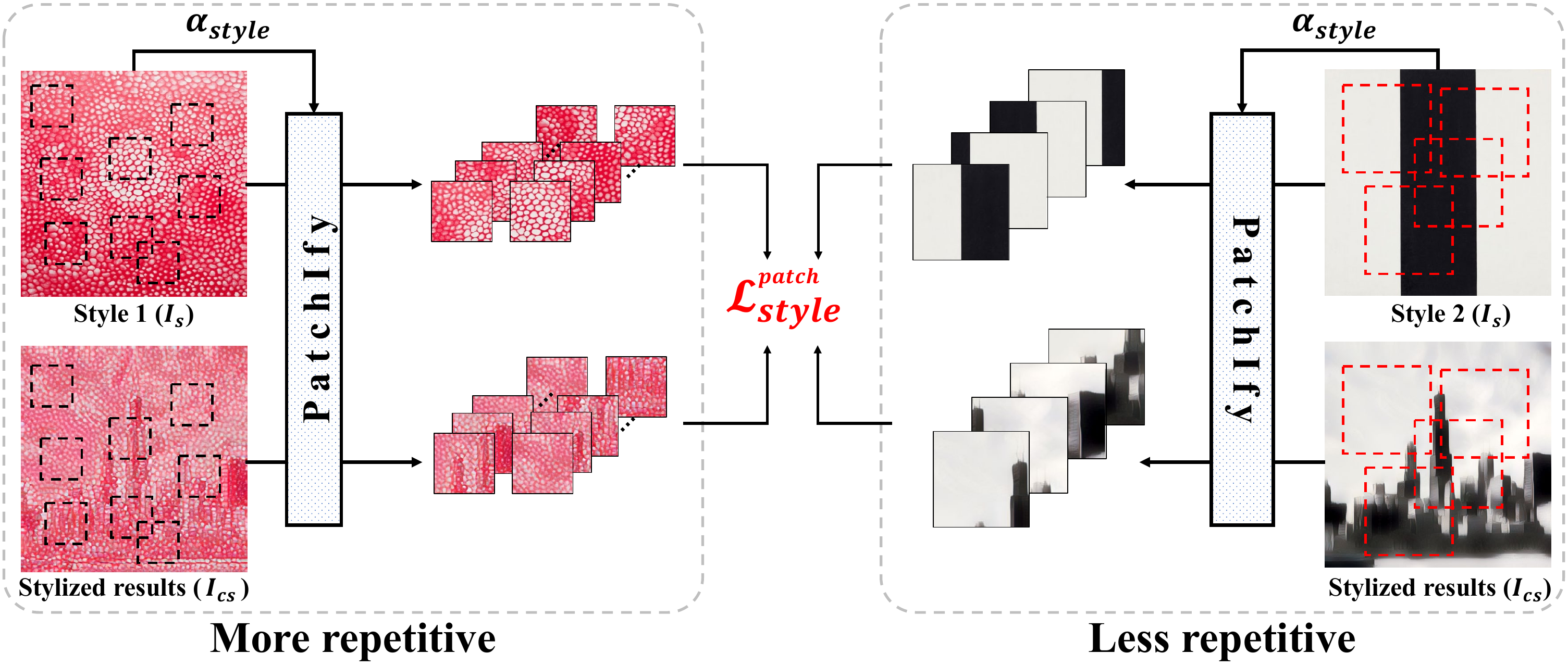}
        \captionof{figure}{The illustration of patch-wise style loss $\mathcal{L}^{patch}_{style}$.}
        \label{fig:patch loss}
        \vspace{-3mm}
\end{figure}

\subsection{Patch-wise Style Loss}
\label{sec:PWstyleloss}
To further convey the aesthetic expression of style into the content, we design the patch-wise style loss ($\mathcal{L}^{patch}_{style}$). Compared to the conventional style loss ($\mathcal{L}^{image}_{style}$) that targets to bring the global style, the proposed patch-wise style loss aims to transfer the elaborate rhythm of local patterns. As shown in \Fref{fig:patch loss}, we patchify the images into $M$ patches $\{\bar{I}_{m}\}_{m=1}^{M}$ by spatially dividing them with the scale $s$, \ie{, $\bar{I}_{m}\ \in \mathbb{R}^{3 \times \frac{H}{s} \times \frac{W}{s}}$}. We define the scale $s$ as $s = max(2^{8\cdot\alpha_{style} - 5}, 1)$ to reflect the pattern repeatability. Afterward, we calculate the $L_2$ distance between the Gram matrices of patches from style and stylized results as follows.
\begin{equation}
\label{equ:patchloss}
    \mathcal{L}^{patch}_{style} =  \frac{1}{M} \sum_{m=1}^{M} \sum_{l=4}^{5} \left\|\mathcal{G}(\phi_{l}(\bar{I}^{cs}_{m})) - \mathcal{G}(\phi_{l}(\bar{I}^{s}_{m}))\right\|_2,
\end{equation}
where $\phi_{l}(\cdot)$ denotes the activation maps after \texttt{ReLU\_}$l$\texttt{\_1} layers of encoder. Note that we utilize the centered-Gram matrix \cite{tmm9143296} which could capture the channel-wise negative correlation. We verify the effectiveness of the proposed loss function in \Sref{sec:stylelosseffect}.

\subsection{Improving Attention Module for AST}
\label{sec:attention} 
We investigate the training procedure of attention mechanisms employed in previous AST methods \cite{park2019arbitrary,Deng2020ArbitraryST}. They train the attention networks to solve the reconstruction task by satisfying the identity loss. However, we insist that this loss design is inappropriate for AST since the content, style, and reconstruction target images are totally identical. As the query features from the content already contain sufficiently rich information for reconstruction, key features from the style are likely to be ignored, thus hindering the attention module from learning useful semantic correspondence between the content and the style.

To alleviate this problem, we introduce an effective self-supervisory task for correspondence learning. We input the original image $I$ and its grayscaled version $I^{gray}$ as the style (key) and the content (query) images, respectively. In this way, due to the reduced amount of information comes from the augmented query, the attention module is required to precisely capture the contextual relationship from reference images. Finally, the loss functions of the self-supervisory reconstruction path are formulated as follows.
\begin{equation}
\label{equ:identity}
    \begin{split}
    \mathcal{L}_{rec} &= \lambda_{identity}\mathcal{L}_{identity}+ \lambda_{cx}\mathcal{L}_{cx}, ~\text{where}\\
    \mathcal{L}_{identity} &= \sum_{l=2}^{5}\left\|\phi_{l}(\tilde{I}) - \phi_{l}(I)\right\|_{2},\\
    \mathcal{L}_{cx} &= CX(\tilde{I}, I),\\
    \end{split}
\end{equation}
where $\tilde{I}$ is the reconstructed image with $I$ and $I^{gray}$. $\phi_{l}(\cdot)$ describes the activation maps after \texttt{ReLU\_}$l$\texttt{\_1} layers of the encoder, \ie, VGG-19. $\mathcal{L}_{cx}$ depicts the contextual similarity loss~\cite{mechrez2018contextual}.

Note that, we modify the reconstruction loss $\mathcal{L}_{rec}$ to be computed with deep features for relieving blurriness and amplifying perceptual quality. The overall reconstruction path is depicted in~\Fref{fig:archi} (a). In \Sref{sec:augmetationeffect} and \ref{sec:stylelosseffect}, we show the validity of the proposed strategy via ablations studies.

\subsection{Training}
As illustrated in~\Fref{fig:archi}, the decoder and the stylization transformer are trained to perform two different tasks (\ie{, reconstruction and stylization}) in an end-to-end manner.
In the reconstruction path, our networks are encouraged to reconstruct the original RGB images when their grayscaled ones are given as the content. Recapitulating the improved attention module in~\Sref{sec:attention}, we induce the stylization transformer to learn semantic correspondence between the content and the style images.

In the stylization path, the style loss for aesthetic style representation is derived by the weighted sum of losses as:
\begin{equation}
\label{eq1:styleloss}
\begin{split}
    \mathcal{L}_{style} &= ~\lambda_{image}\mathcal{L}^{image}_{style} + \lambda_{patch}\mathcal{L}^{patch}_{style} +  
    \lambda_{lf}\mathcal{L}_{lf} \\&~~+ \lambda_{content}\mathcal{L}_{content} + \lambda_{color}\mathcal{L}_{color} + \lambda_{tv}\mathcal{L}_{tv}.
\end{split}
\end{equation}
Notably, the style loss $\mathcal{L}^{image}_{style}$ and the proposed patch-wise style loss $\mathcal{L}^{patch}_{style}$ contribute to global and local target texture expression, respectively. Besides, the content loss $\mathcal{L}_{content}$ is adopted for the content as well as context preservation. In addition, we adopt the local feature loss $\mathcal{L}_{lf}$ \cite{Liu2021AdaAttNRA} to encourage feature consistency. Moreover, we utilize the color loss $\mathcal{L}_{color}$ \cite{Afifi2021HistoGANCC} to follow the color palettes of the style image, and the total variation loss $\mathcal{L}_{tv}$ to encourage pixel-wise smoothness. Each loss is described as follows.
\begin{equation}
\label{eq1:detail loss}
\begin{split}
    \mathcal{L}^{image}_{style} &= \sum_{l=4}^{5}\left\|\mathcal{G}(\phi_{l}(I^{cs})) - \mathcal{G}(\phi_{l}(I^{s}))\right\|,\\
    \mathcal{L}_{content} &= \sum_{l=4}^{5}\left\|\phi_{l}(I^{cs}) - \phi_{l}(I^{c})\right\|,\\
    \mathcal{L}_{lf} &= \sum_{l=4}^{5}\left\|\phi_{l}(I^{cs}) - f^{cs}_{attn} \right\|,\\
    \mathcal{L}_{color} &= \frac{1}{\sqrt{2}}\left \| (H^{s})^{1/2} - (H^{cs})^{1/2} \right \|,
\end{split}
\end{equation}
where $H^{1/2}$ denotes the element-wise square root of the color histogram. To further elevate the stylization performance, we employ the multi-scale discriminator with the adversarial loss (${\mathcal{L}_{adv}}$) \cite{Hong2021DomainAwareUS}.

Finally, the full objective of our model is summation of those loss functions: $\mathcal{L} = \mathcal{L}_{rec} + \mathcal{L}_{style} +\lambda_{adv}\mathcal{L}_{adv}$.

\section{Experiments}
\noindent\textbf{Implementation details.}~~We train AesPA-Net on MS-COCO \cite{lin2014microsoft} and WikiART \cite{phillips2011wiki} datasets, each containing about 120K real photos and 80K artistic images. We use the Adam~\cite{Kingma2015AdamAM} optimizer to train the attention module as well as the decoder with a batch size of 6, and the learning rate is set to $1e$-4. During the training and inference, we rescale the input images to have a spatial resolution of 256$\times$256. We set the weighting factors of loss functions as follows: $\lambda_{identity}=1$, $\lambda_{cx}=1$, $\lambda_{content}=1$, $\lambda_{image}=10$, $\lambda_{lf}=100$, $\lambda_{patch}=0.5$, $\lambda_{color}=1$, $\lambda_{adv}=0.1$, $\lambda_{tv}=1$. We implement our framework with PyTorch~\cite{paszke2017automatic} and train our model using a single GTX 3090Ti.

\subsection{Analysis}
\subsubsection{Reliability of the pattern repeatability}
\label{sec:analysis}
\begin{figure}[t]
        \centering
        \includegraphics[width=1.0\linewidth]{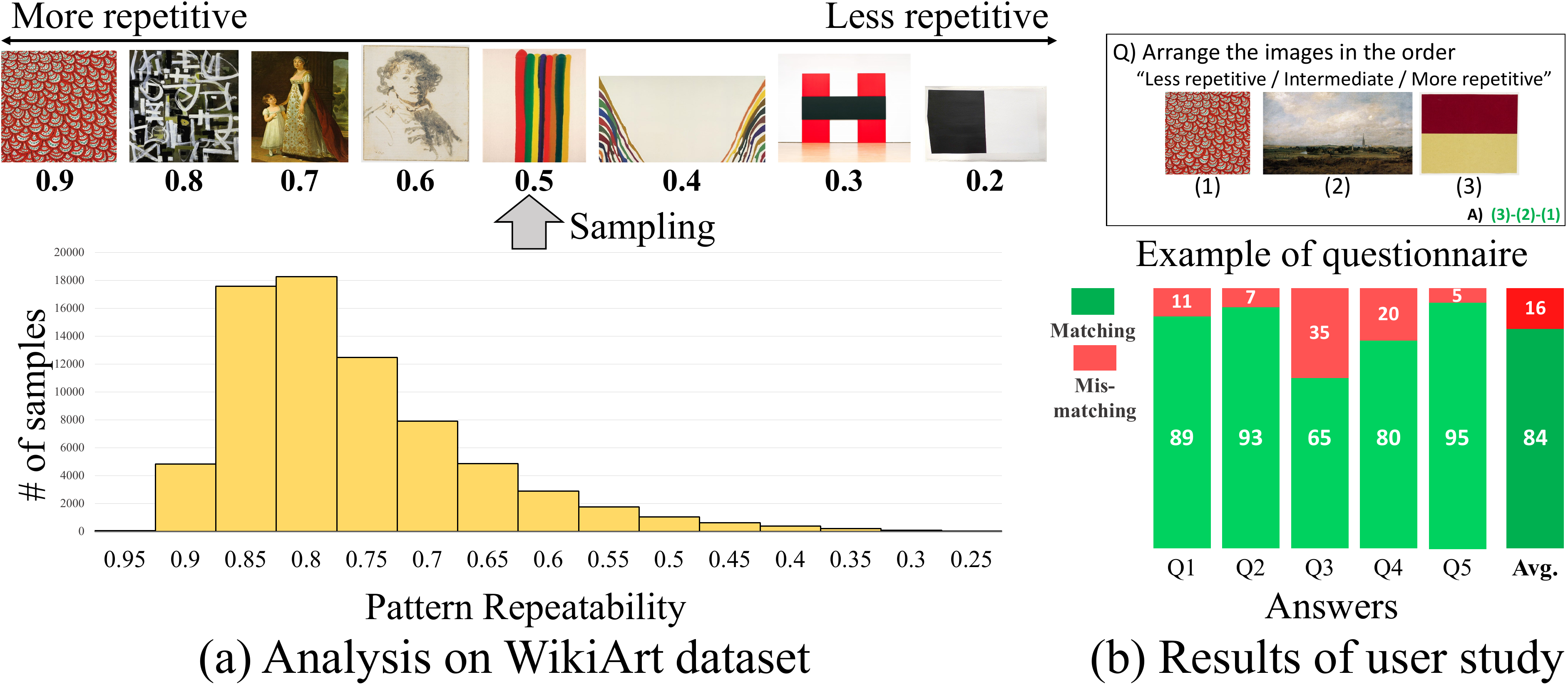}
        \captionof{figure}{Analysis of the proposed pattern repeatability $\alpha_{style}$. (a) presents the distribution of WikiArt \cite{phillips2011wiki} dataset and examples according to $\alpha_{style}$. (b) shows the results of user study on correlation of $\alpha_{style}$ with the human perception.}
        \label{fig:analysis}
        \vspace{-2mm}
\end{figure}

We analyze the proposed pattern repeatability ($\alpha_{style}$) in order to validate its reliability. Firstly, we analyze the WikiArt \cite{phillips2011wiki} dataset by computing the pattern repeatability of each artwork to examine whether the extracted $\alpha_{style}$ matches well with the visual pattern rhythm of the image. We depict the histogram of pattern repeatability in \Fref{fig:analysis} (a). The mean of pattern repeatability is about 0.79 and the standard deviation is 0.1. The highest and lowest value of $\alpha_{style}$ is 0.97 and 0.26, respectively. 

For qualitative verification of pattern repeatability, we visualize randomly sampled images for every $0.1$ range. As shown in the upper part of \Fref{fig:analysis} (a), images with high $\alpha_{style}$ have a uniformly repeated local pattern, and the local patterns thus can express the global pattern regardless of its scale. On the other hand, images with low $\alpha_{style}$ tend to have irregular local patterns, and their style hence should be interpreted from the entire image.

\begin{figure}[t]
        \centering
        \includegraphics[width=1.0\linewidth]{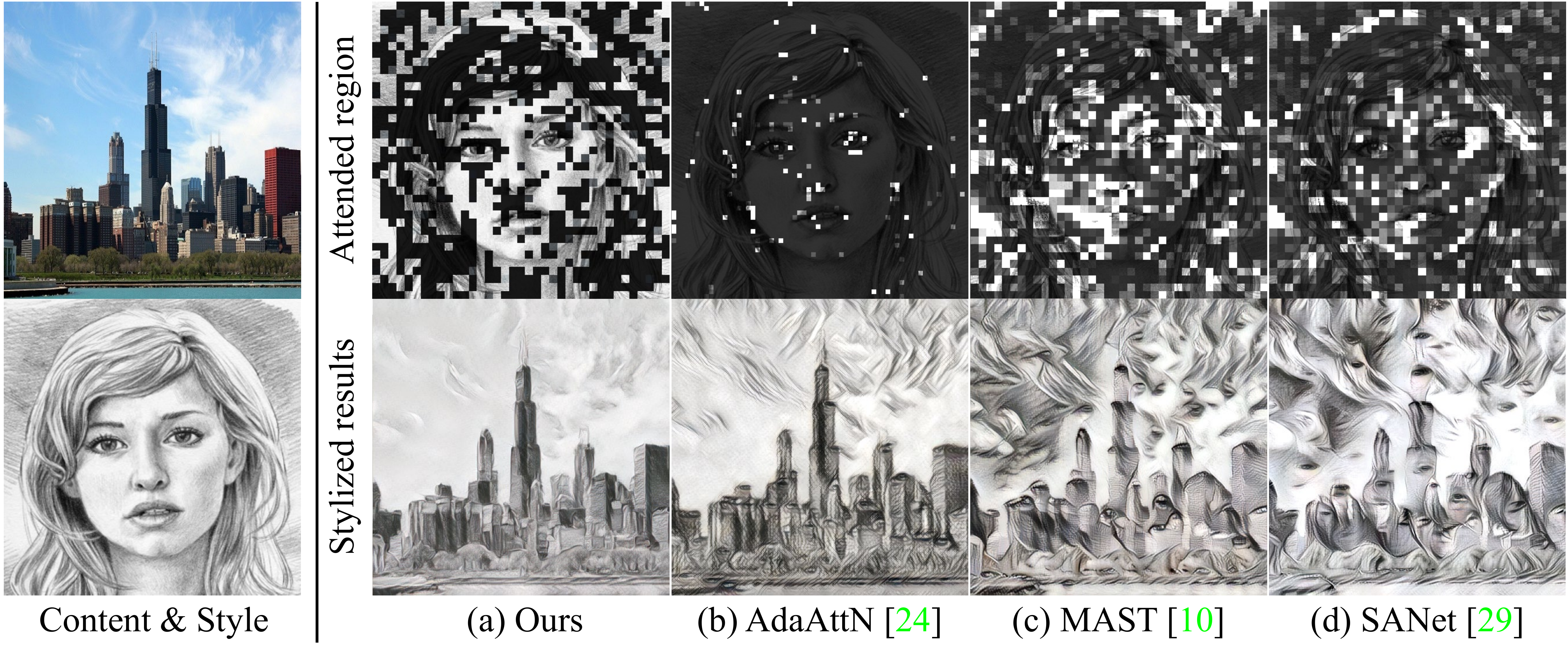}
        \captionof{figure}{Visualization of attended region (first row) and stylized results (second row) given a pair of content and style images.}
        \label{fig:suppleattend}
        \vspace{-2mm}
\end{figure}

\begin{figure*}[t]
        \centering
        \includegraphics[width=1.0\linewidth]{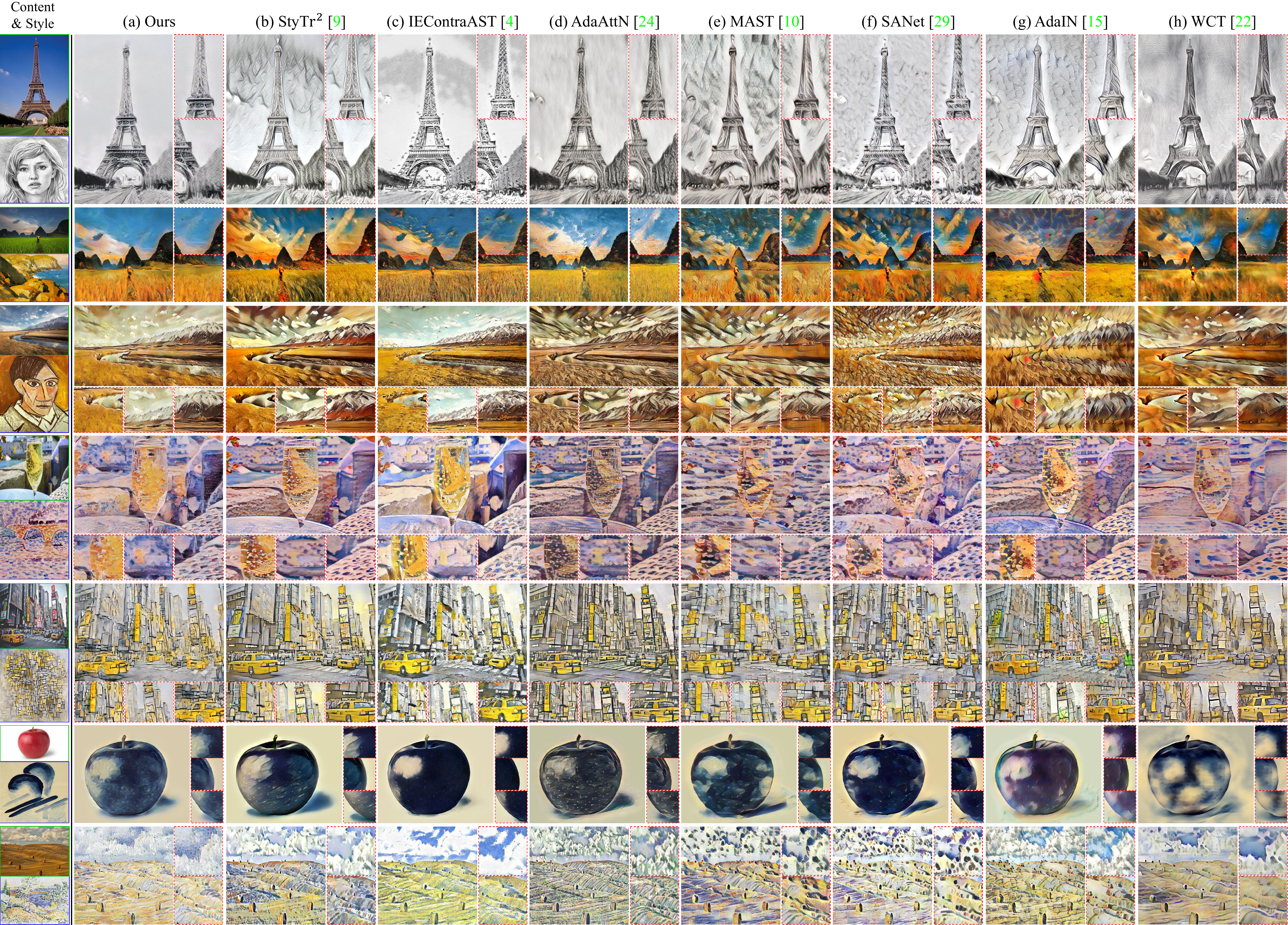}
        \captionof{figure}{Qualitative comparisons with state-of-the-art AST methods. \textcolor{green}{Green} and  \textcolor{blue}{blue} boxes indicate the content and reference images respectively. We provide zoomed patches (\textcolor{red}{dashed red} boxes) for better comparison.}
        \label{fig:qual}
\end{figure*}

To further demonstrate that the proposed pattern repeatability aligns well with human perception, we conduct the user study. We compose each questionnaire with three randomly shuffled images of different pattern repeatability $\alpha_{style}$ from ``Less repetitive" ($0.1$$\sim$$0.3$) to ``More repetitive" ($0.8$$\sim$$0.9$). Then, we asked the participants to arrange the images in ascending order to compare the responses with the sequence according to the proposed metric. In \Fref{fig:analysis} (b), the answers mostly agree with the sequence of the proposed metric, reporting the average matching rate of 84\%. This verifies that $\alpha_{style}$ reflects the pattern repetition as the perception of the human vision system.

\subsubsection{Effectiveness of improved attention module}
\label{sec:augmetationeffect}
We demonstrate the effectiveness of the improved attention module compared to the previous attention-based baselines \cite{park2019arbitrary,Deng2020ArbitraryST,Liu2021AdaAttNRA}. The proposed self-supervisory task reinforces the overall style transmission effect. As shown in \Fref{fig:suppleattend}, the attention module from previous works \cite{park2019arbitrary,Deng2020ArbitraryST,Liu2021AdaAttNRA} selects some limited local patches (\eg{, eyes}), leading to disharmonious artifacts in stylized results. On the other hand, our proposed grayscaled augmentation strategy helps the attention module to capture meaningful and broader correspondence between the content and the style image. Finally, AesPA-Net generates plausible artistic outputs without disharmonious artifacts.

\subsection{Qualitative Results}
As shown in \Fref{fig:qual}, we qualitatively compare our method with five state-of-the-art attention-based AST models \cite{deng2022stytr2,Chen2021ArtisticST,Liu2021AdaAttNRA,Deng2020ArbitraryST,park2019arbitrary} and two global statistic-based methods \cite{huang2017arbitrary,li2017universal}. Specifically in the 1$^{st}$ and 2$^{nd}$ rows, most works \cite{deng2022stytr2,Liu2021AdaAttNRA,Deng2020ArbitraryST,park2019arbitrary,huang2017arbitrary} generate repetitive artifacts in stylized results, \eg, wave stains on the sky region. Although these artifacts are rarely appearing in IEContraAST \cite{Chen2021ArtisticST}, it still produces disharmonious distortions of eyes around the Eiffel tower and struggles to transfer detailed texture and strokes of \textit{Van Gogh}'s painting. On the other hand, AesPA-Net successfully reproduces the contents respecting the texture and color palette of the artworks without artifacts.

\begin{table*}[t]
\begin{center}
\resizebox{0.95\linewidth}{!}
{
\begin{tabular}{c|cccccccc}
\cline{1-9}\cline{1-9}
 & AesPA-Net~(Ours)~ & StyTr$^{2}$ \cite{deng2022stytr2} & IEContraAST \cite{Chen2021ArtisticST} & AdaAttN \cite{Liu2021AdaAttNRA} & MAST \cite{Deng2020ArbitraryST} & SANet \cite{park2019arbitrary}  & AdaIN \cite{huang2017arbitrary}  & WCT \cite{li2017universal}\\
\hline\hline

User preference &  0.6$5^{*}$  & \textbf{0.65} / 0.35 & \textbf{0.70} / 0.30 & \textbf{0.67} / 0.33 & \textbf{0.58} / 0.42 & \textbf{0.68} / 0.32 & \textbf{0.57} / 0.43 & \textbf{0.72} / 0.28
\\

Human deception rate&  0.7$2^{*}$  & \textbf{0.63} / 0.37 & \textbf{0.68} / 0.32 & \textbf{0.63} / 0.37 & \textbf{0.77} / 0.23 & \textbf{0.74} / 0.26 & \textbf{0.70} / 0.30 & \textbf{0.87} / 0.13   \\
\cline{1-9}
Content fidelity ($\uparrow$)\cite{Wang2021EvaluateAI} &  \underline{0.690} & 0.685 & \textbf{0.698} & 0.645 & 0.554 & 0.567 & 0.609 & 0.437    \\
Style loss ($\downarrow$) \cite{gatys2016image} &  \textbf{0.258} & 0.305 & 0.316 & 0.278 & \underline{0.274} & 0.313 & 0.284 & 0.345    \\
Pattern difference ($\downarrow$) &  \textbf{0.075} & 0.109 & 0.113 & 0.100 & 0.104 & 0.097 & \underline{0.091} & 0.109    \\
Inference time (Sec.) & 0.065  & 0.445 & 0.381 & 0.054 & 0.020 & 0.015 & 0.349 & 1.071    \\
\cline{1-9}\cline{1-9}
\end{tabular}}
\end{center}
\vspace{-2mm}
\caption{Quantitative comparisons with state-of-the-art AST methods. * denotes the average user responses.}
\label{tab:quant}
\vspace{-2mm}
\end{table*}


With style images of high pattern repeatability (3$^{rd}$, 4$^{th}$, and 5$^{th}$ rows), our proposed methods infuse the delicate rhythm of target patterns into the content images. Particularly, the glass surface is expertly brushed with vivid yellow strokes of the bridge, and the urbanscape is adeptly illustrated with \textit{Mondrian}'s signature irregular rectangles. Moreover, the hills and sky are represented through skillful stippling. However, some methods produce photo-realistic drawings (b-d) or fall short of representing local patterns of the style image (e-h).

Lastly, with less-repetitive styles  (6$^{th}$ and 7$^{th}$ rows), the proposed method effectively describes the global representation without mottled artifacts. Notably, patches of the stylized apple most closely resemble the style of a given single brushstroke drawing. To summarize, AesPA-Net synthesizes artistic stylized results with high fidelity in terms of texture, color, and pattern repeatability across a wide range of style patterns. We provide additional qualitative comparisons in the supplementary material.

\subsection{Quantitative Results}
\noindent\textbf{User study.} Since judging successful artistic stylization is quite subjective, we conducted a user study to compare the quality of artistic expression (\ie{, user preference}) and the reality of stylized results (\ie{, human deception rate}). We sampled 42 stylized results from random pairs of content and style images, and conduct one-to-one comparisons with representative AST models \cite{park2019arbitrary,Deng2020ArbitraryST,Liu2021AdaAttNRA,Chen2021ArtisticST,deng2022stytr2,li2017universal,huang2017arbitrary}. We collected 2310 responses from 55 subjects.

As reported in the 1$^{st}$ and 2$^{nd}$ rows of \Tref{tab:quant}, most participants favored our AesPA-Net over competitors in terms of artistic representation. Plus, when participants are asked to choose more realistic artwork following previous works \cite{Sanakoyeu2018ASC,Kotovenko2019ACT}, AesPA-Net surpassed previous methods with the evident gap in terms of the human deception rate.

\noindent\textbf{Statistics.} Following previous works~\cite{wang2022aesust,Wang2021EvaluateAI,Hong2021DomainAwareUS}, we respectively compute degrees of artistic stylization and contextual preservation of content information via two proxy metrics: style loss \cite{gatys2016image} and content fidelity (CF) \cite{Wang2021EvaluateAI, wang2022aesust}. To calculate CF, we estimate the cosine-similarity between features of content and stylized images as follows: $CF(I^{cs}, I^{c}) =\frac{1}{L}\sum^{L}_{l=1} \frac{f_{l}^{cs}\cdot f_{l}^{c}}{{\left\|f_{l}^{cs}\right\|}{\left\|f_{l}^{c}\right\|}},$ where $f_{l}^{c}$ and $f_{l}^{cs}$ respectively are content and stylized features extracted from $l^{th}$ convolutional layer of an encoder. Additionally, we compute the style loss \cite{gatys2016image} between artistic stylized outputs and reference samples over five crop patches, following DSTN \cite{Hong2021DomainAwareUS}. We sample 65 random pairs of content and style references for quantitative evaluation.

The 3$^{rd}$ and 4$^{th}$ rows of \Tref{tab:quant} depict the quantitative results compared with previous AST methods. Noticeably, they fall behind either in preserving the original context (\eg, WCT \cite{li2017universal}, SANet \cite{park2019arbitrary}, MAST \cite{Deng2020ArbitraryST}), or in expressing artistic styles (\eg,  IEcontraAST \cite{Chen2021ArtisticST}, StyTr$^{2}$ \cite{deng2022stytr2}). In contrast, AesPA-Net achieves state-of-the-art results in terms of the style loss while being competent in the content fidelity. To further investigate the scalability of our proposed metric, \textit{pattern repeatability}, we assess the quality of AST by quantifying the $L1$-distance of pattern repeatability between reference and stylized images as follows: $||\alpha_{style}-\alpha_{stylization}||_{1}$. As shown in the 5$^{th}$ arrow of \Tref{tab:quant}, AesPA-Net transfers the specific pattern well and the pattern repeatability could be a new metric for AST. 

\noindent\textbf{Inference time.} We record the inference time with a input resolution of 512$\times$512 compared with other methods in the last row of \Tref{tab:quant}. The inference speed of our AesPA-Net is comparable to state-of-the-art methods, and therefore it can be adopted in practice for real-time image stylization. 
\begin{figure}[t]
        \centering
        \includegraphics[width=1.0\linewidth]{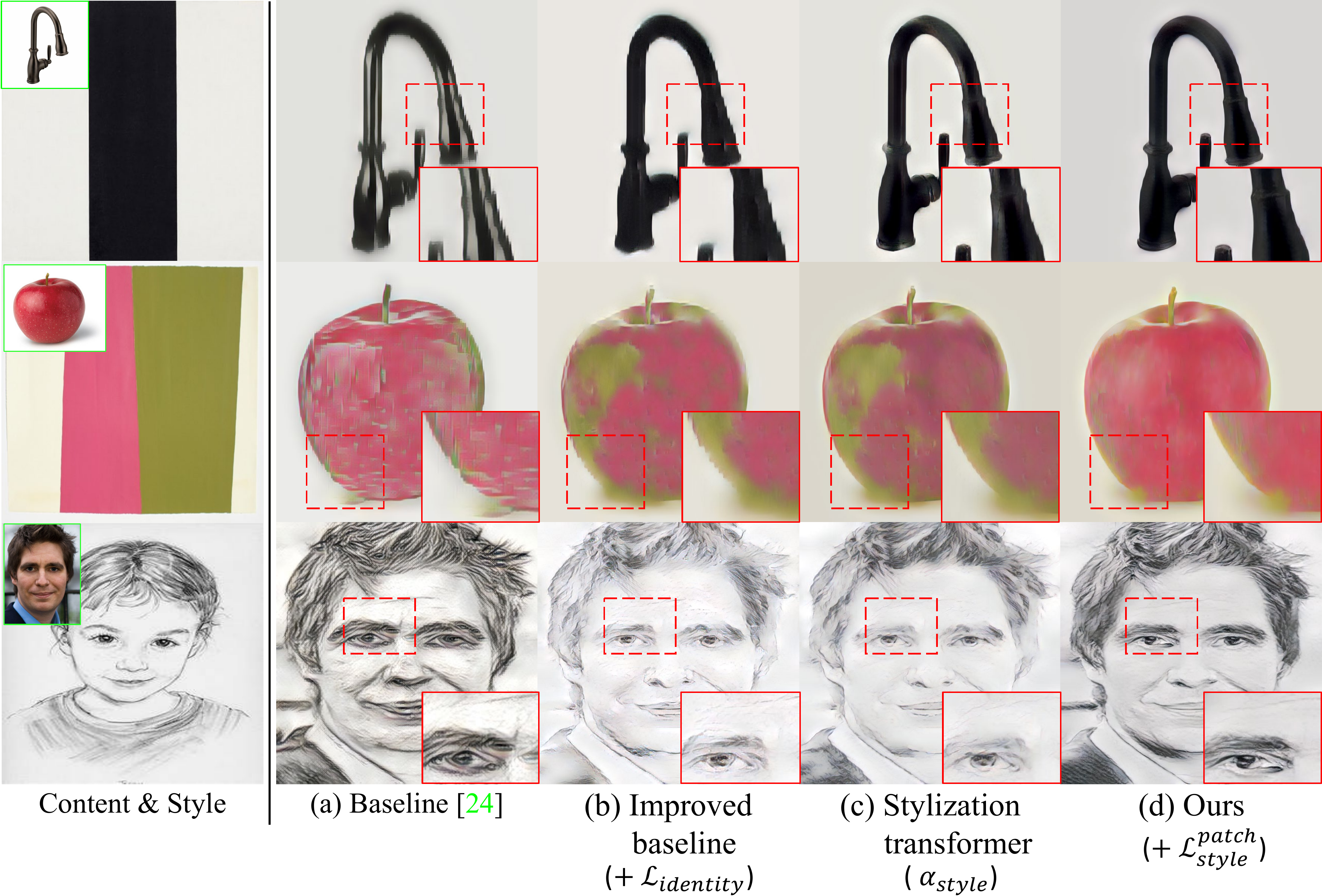}
        \captionof{figure}{Qualitative ablation studies of AesPA-Net. From the baseline (a) to the full model (d), we show the effectiveness of proposed components sequentially. The \textcolor{green}{green box} indicates the input content image and \textcolor{red}{red box} depicts the zoomed patches.}
        \label{fig:ablation}
        \vspace{-2mm}
\end{figure}

\subsection{Ablation Studies}
\label{sec:stylelosseffect}
In \Fref{fig:ablation}, we demonstrate the effectiveness of each proposed component of AesPA-Net qualitatively: the improved attention module training (b), stylization transformer (c), and the patch-wise style loss (d). Compared to the attention-based baseline \cite{Liu2021AdaAttNRA}, our proposed self-supervisory reconstruction task helps the attention module to capture meaningful and broader correspondence regions from the style image. For example in the first row, the baseline fills the texture of a water tap with the stripe pattern, whereas the improved baseline paints the water tap with black as the style image. Also in the last row, artifacts of repetitive eye patterns are removed in (b). As the proposed grayscale augmentation strategy guides the attention module to find dense semantic correspondence, the local textures in the style image are delivered to the proper regions in the content.

Besides, the proposed stylization transformer removes defective artifacts caused by unnatural mapping from the attention module. Particularly, the aliased outline of a water tap is smoothed and the noisy patterns in the apple are cleared out. As the stylization transformer adaptively fuses the stylized features $f_{attn}^{cs}$ and $f_{global}^{cs}$ in accordance with the $\alpha_{style}$, it relieves the artifacts by reflecting the global style pattern. Also, the features from the attention module $f_{attn}^{cs}$ prevent the exaggerated distortion of the original context.

Lastly, the patch-wise style loss boosts the overall style expression and reduces hallucinations in the skin, as shown in \Fref{fig:ablation} (d). Since we utilize $\alpha_{style}$ to find the minimal patch $\bar{I}^{s}_{m}$ that can express the global style image by its repetition, its feature gram matrix $\mathcal{G}(\phi_{l}(\bar{I}^{s}_{m}))$ represents a unit style of the global pattern rhythm. Therefore, exploiting it as a guideline results in delicate and aesthetic stylization.

\section{Conclusion}
In this paper, we revisit the definition of style with pattern repeatability. The proposed pattern repeatability quantifies the rhythm and repetition of local patterns in the style images by taking into account inter-patch similarity and intra-image patch similarity. Based on the pattern repeatability, we designed a novel style transfer framework, \textit{AesPA-Net}, that can deliver the full extent of the target style via the stylization transformer. Extensive experiments show that our AesPA-Net successfully conducts elaborate artistic style transfer with various style patterns. Our future goal is to expand our method for more impeccable artistic style transfer by considering not only repeatability but also the context in the style image, such as relative locations of style patches.\\

\vspace{-2mm}
\noindent\textbf{Acknowledgements.} 
{\small This project was supported by the National Research Foundation of Korea grant funded by the Korean government (MSIT) (No. 2022R1A2B5B02001467).}

{\small
\bibliographystyle{ieee_fullname}
\bibliography{final}
}

\end{document}